\documentclass[12pt]{article}

\usepackage{sbc-template}
\usepackage{booktabs}
\usepackage{hyperref}

\usepackage{graphicx,url}

\usepackage[utf8]{inputenc}  

\title{DEBISS: a Corpus of Individual, Semi-structured and Spoken Debates\thanks{A shortened version of this paper was published in the proceedings of STIL 2025. If you wish to cite our work, please refer to the published version: \url{https://doi.org/10.5753/stil.2025.37860}. You may also wish to cite the following related works: [\url{https://doi.org/10.5753/stil.2025.37836}, \url{https://doi.org/10.5753/jbcs.2025.5824}].}}

\author{Klaywert Danillo Ferreira de Souza\inst{1}, David Eduardo Pereira\inst{1}, \\ Cláudio E. C. Campelo\inst{1},  
Larissa Lucena Vasconcelos\inst{2} }

\address{Federal University of Campina Grande (UFCG) - System and Computing
\\ 
Campina Grande, Brazil
\nextinstitute
  Federal Institute of Paraíba (IFPB) \\
  Monteiro, Brazil
  \email{klaywertdanillo@copin.ufcg.edu.br  , david.pereira@ccc.ufcg.edu.br}
  \email{campelo@dsc.ufcg.edu.br, larissalucena@gmail.com}
}
\begin{document} 

\maketitle

\begin{abstract}
The process of debating is essential in our daily lives, whether in studying, work activities, simple everyday discussions, political debates on TV, or online discussions on social networks. The range of uses for debates is broad. Due to the diverse applications, structures, and formats of debates, developing corpora that account for these variations can be challenging, and the scarcity of debate corpora in the state of the art is notable. For this reason, the current research proposes the DEBISS corpus — a collection of spoken and individual debates with semi-structured features. With a broad range of NLP task annotations, such as speech-to-text, speaker diarization, argument mining, and debater quality assessment.
\end{abstract}

\section{Introduction}

Debates are pivotal in education, significantly enhancing critical thinking and communication skills. Despite their importance, progress in the computational analysis of debate discourse is constrained by a scarcity of diverse, richly annotated corpora. This limitation restricts research avenues and the development of specialized computational models tailored to nuanced debate interactions.

Moreover, there is a particular dearth of resources for oral, semi-structured debates within educational contexts, especially for languages other than English. To fill this gap, we introduce DEBISS (Spoken, Individual and Semi-structured Debates), a novel annotated Brazilian Portuguese corpus. It captures 9 hours and 35 minutes of in-person, individual, semi-structured debates centered on ‘Generative Artificial Intelligence and its impacts on society,’ conducted among 67 first-semester computer science students from Federal University of Campina Grande.

Hence, DEBISS corpus offers a unique dataset by focusing on spoken, individual debates that feature a blend of predefined questions and moments for open expression, allowing for spontaneous argumentation distinct from highly structured or written formats. Key contributions include its focus on Brazilian Portuguese, enriching language diversity in debate analysis, and its grounding in an educational setting, providing insights into student oratory skill development. The corpus is further enhanced by detailed self and peer evaluations from participants regarding performance and topic knowledge, alongside multimodal data (audio and transcriptions) crucial for comprehensive analysis and its contemporary theme ensures relevance.

This comprehensive corpus facilitates exploration of argumentative discourse, from argument construction and rhetorical strategies to debater interaction patterns, and is valuable for NLP (Natural Language Processing) tasks such as Argument Mining (AM) and Debater Quality Analysis (DQA), speech to text, speaker diarization, disfluency detection and so on. The DEBISS corpus, including all transcriptions, audio files, and associated annotations, will be made available for research purposes via GitHub\footnote{https://github.com/AINDA-Project-UFCG/DEBISS}.

The rest of this article is structured as follows: Section 2 reviews related research on debate corpora and examines the knowledge representation and methodology adopted for creating datasets in this field. Section 3 details the methodology for data collection and processing, corpus analysis and statistical insights. Finally, conclusions on the dataset are presented in Section 4.

\section{Related Work}

The study and analysis of debates have attracted significant attention across various fields, including computational linguistics, discourse analysis, and artificial intelligence. Researchers have developed and utilized numerous corpora to understand and model different aspects of argumentative discourse. This section reviews prominent debate corpora and methodologies, focusing on their structures, annotation schemes, and applicability to computational tasks such as Argument Mining (AM) and Debater Quality Analysis (DQA).

\subsection{Political Debate Corpora}

Political debates have been a primary focus for constructing argumentative discourse corpora due to their structured nature and public availability. Notable among these is the U.S. Presidential Debate Corpus \cite{vrana2017saying}, and other political sources \cite{0eac738616614094950bb74635ce3d49, mestre-etal-2021-arg, mancini-etal-2022-multimodal}, which comprises transcripts from various presidential debates annotated for argumentative structures and rhetorical strategies \cite{de2018polly, carvalho2011liars, hautli2022qt30}. Despite their utility, political debate corpora often exhibit high levels of formality and adherence to strict protocols, which may not reflect the dynamics of less structured or spontaneous argumentative interactions. Additionally, the focus on specific contexts and topics limits the generalization of models trained on these datasets to other forms of debate and discourse.

\subsection{Online Debate and Discussion Corpora}

The proliferation of online platforms has led to the creation of corpora derived from digital discussions and debates \cite{durmus2019corpus, khodak2017large, stranisci2021expression, lai2018stance}. Researches related to debates deal with written debates on online platforms \cite{brasnam, habernal-gurevych-2016-argument, boltuzic-snajder-2016-fill, chakrabarty-etal-2019-ampersand}, such as Twitter and Reddit. 

The Internet Argument Corpus (IAC) \cite{abbott2016internet} aggregates discussions from platforms like Reddit and Forums, annotated for argument stance, quality, and relevance. This corpus captures a wide range of topics and styles, providing a rich resource for studying informal argumentative discourse and sentiment analysis. Online debate corpora often suffer from issues related to noise, informal language, and varying discourse structures, posing challenges for consistent annotation and analysis. Moreover, these datasets predominantly represent written communication, lacking the multimodal and spontaneous elements present in spoken debates.

\subsection{Educational and Academic Debate Corpora}

Academic settings provide a fertile ground for studying argumentative skills development and discourse proficiency. Most educational corpora \cite{ruiz2021vivesdebate} includes recordings and transcripts of academic formal debates, offering insights into educational discourse but not specifically focusing on less structured debates. Existing educational corpora often do not focus explicitly on debates, especially those that are semi-structured and involve spontaneous argumentation. There is a paucity of annotated datasets capturing the nuances of student-led debates, particularly in languages other than English, limiting cross-cultural and multilingual studies in argumentative discourse analysis.

\section{Methodology}

Debates are a form of human interaction that occur regularly in daily life and have evolved over time as a fundamental aspect of communication. This research aims to construct a debate corpus with well-defined characteristics. To accomplish this objective, a methodology for data collection and processing was developed, following specific guidelines throughout the corpus creation process. This section provides a detailed explanation of the methodological steps designed and implemented in the development of the proposed corpus.

\subsection{DEBISS Corpus}

The DEBISS corpus comprises audio transcriptions of in-person debates conducted with the consent of first-year computer science undergraduate students at (Omitted) University. Each debate session was moderated by a facilitator. Data collection took place in 2024 and involved \textbf{67} students, who were organized into \textbf{16} debate groups, generating a total of \textbf{9 hours} and \textbf{35 minutes} of audio recordings. These recordings were transcribed using a semi-automated process that combined speech-to-text AI models with human validation to ensure accuracy.

Moreover, debate formats may vary across corpora, some adopt a monologue-based style, others feature individual debaters defending personal viewpoints, and some use group-based approaches where a single perspective is collectively defended. In this study, debate groups consisted of \textbf{3} to \textbf{5} participants, totaling 16 groups. Each participant defended their own viewpoint, speaking independently. This individual-focused format is consistent with methodologies employed in other corpora that emphasize each debater personal argumentative characteristics.

\subsubsection{Debate Subject}

To create a focused and engaging debate, a central theme was selected: ``Generative Artificial Intelligence and Its Impacts on Society''. This topic is highly relevant and controversial, prompting extensive questioning, criticism, across various fields. Defining a central theme is common in some corpus methodologies to guide debates. However, some corpora, particularly those focused on political debates, do not limit themselves to a specific theme, as multiple aspects of the political landscape — such as security, international relations, healthcare, and education — can be discussed.

Furthermore, to stimulate the debate process, it was compiled a collection of online texts to be shared with the debaters ahead of the debate. These texts were selected for their readability, prioritizing news articles and opinion pieces from reputable sources over complex academic papers. The chosen texts addressed specific, relevant, and controversial issues related to the theme, such as the impact of AI on jobs, the use of generative AI in education, and the legal implications of AI solutions. Since reading the selected texts was voluntary, the research team also provided a two-page summary, termed the ``supplementary text''. This summary condensed key information on critical topics.

\subsubsection{Debate Environment Setup}

To collect detailed data from the debates, the sessions were recorded using a Logitech USB Yeti Condenser Microphone in omnidirectional mode, paired with OBS Studio software. The audio was captured in a 3x5m conference room where debaters sat around a central table. OBS was configured to the optimal decibel level for all participants, producing MKV files that were later converted to MP3 format. The room also included a 55-inch TV displaying debate information, and the layout is depicted in an accompanying Figure \ref{fig:exampleFig1}.

\begin{figure}[ht]
\centering
\includegraphics[width=.6\textwidth]{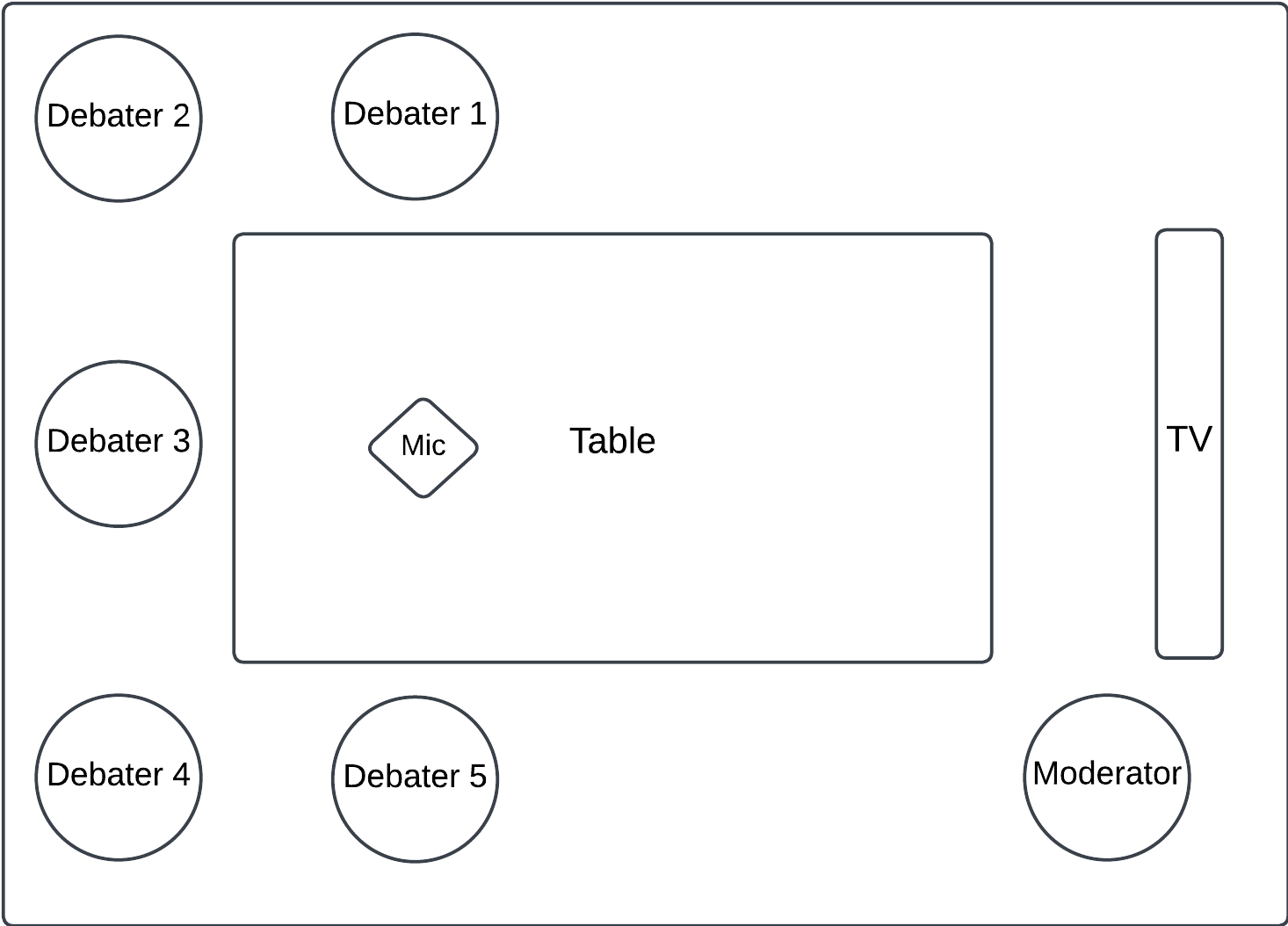}
\caption{Room layout during debate sessions}
\label{fig:exampleFig1}
\end{figure}

\subsubsection{Debate Format}

The data collection protocol was designed to capture speech data through semi-structured debates. All debate sessions adhered to a standardized semi-structured format to ensure consistency throughout the data-gathering process. The session begins with an explanation of the study’s purpose, informing the voluntary participants that the debate will be recorded without including any personal information. The primary goal is to create a transcribed debate corpus for use in various scenarios, particularly in developing AI models focused on debate analysis. The detailed structure of these sessions is outlined below:

\begin{enumerate}
    \item Presentation and explanation of the motivation for the corpus creation
    \item Signature of the voice and image consent form
    \item Assignment of unique numeric identifiers to each debater
    \item Recording of the debaters' voice samples for identification
    \item Explanation of the debate rules
    \item Initial stance and thoughts on the topic
    \item Question and answer rounds
    \item Final stance and reflections on the topic
    \item End recording
    \item Completion of the self-evaluation form
\end{enumerate}

Next, participants are then asked to sign a consent form authorizing the use of their recorded voices for research purposes. This step is essential, and the research was formally approved by the university’s ethical committee due to the involvement of human subjects. To aid in voice identification during manual analysis, each debater initially records the same sentence before the debate begins. With signed consent and voice samples collected, the debate session proceeds. A moderator oversees the debate, ensuring order and encouraging discussion. The moderator first explains the rules: debaters should listen to the moderator, refrain from interrupting each other, and raise their hand if they wish to speak, waiting for their turn or for the current speaker to finish.

Also, the debate section is divided into three parts. The \textbf{first part} allows debaters to express their initial opinions on the topic. The \textbf{second part} consists of a question-and-answer round, and the \textbf{third part} involves final thoughts and reflections on the topic. Each debater is assigned specific questions, which are read aloud by the moderator and displayed on a TV. After answering, other debaters may comment or ask additional questions. This interaction is optional, providing an open space for engagement while maintaining a semi-structured format with both mandatory and voluntary interactions. Each debater has uninterrupted time to express their opinions and answer questions. A total of five questions are prepared, with one question per debater; if there are only three participants, only the first three questions are used.

After the question-and-answer segment, there is a final question for all participants, allowing those who wish to contribute additional insights. This interaction is also optional, fostering a rich and participatory debate that encourages the exchange of diverse ideas and perspectives.  In the final part of the debate, participants are asked to share any concluding thoughts and whether their opinions have changed based on the discussion. This final reflection is mandatory, giving each debater an opportunity to express their final stance.

\subsection{Debater Evaluation}

Once the debate and recording ended, each participant completed a self-evaluation form (Google Forms). This was crucial for assessing individual performance and personal reflections on the debate process, providing valuable data for analysis. The questions were designed to capture insights into the debater's preparation, engagement, and overall experience. Identification numbers ensured that responses could be linked to the debate data while maintaining confidentiality. The self-evaluation form included the following questions:

\begin{itemize}
    \item \textit{What is your identification number?} (Multiple choice)
    \item \textit{How would you rate your performance during the debate?} (Likert Scale: Poor to Excellent)
    \item \textit{Before the debate, how well did you know the topic (from study or experience)?} (Likert Scale: Strongly Disagree to Strongly Agree)
    \item \textit{Did you thoroughly read the material provided by the organizers? }(Likert Scale: Strongly Disagree to Strongly Agree)
    \item \textit{Did you study additional materials beyond what was provided?} (Likert Scale: Strongly Disagree to Strongly Agree)
    \item \textit{Did the debate expand your perspective or knowledge on the topic? }(Likert Scale: Strongly Disagree to Strongly Agree)
    \item \textit{Who do you think was the best debater?} (Multiple Choice, including the option "there was no best debater")
    \item \textit{Justify your choice for the best debater.} (Open Text)
\end{itemize}

Participants rated their own performance, reflecting on their effectiveness in argumentation, which provided insights into their self-perception within the debate dynamics. Questions about prior knowledge and preparation assessed the participants' familiarity with the topic and their engagement with the provided materials. This helped to assess how preparation influenced their performance. The question on whether the debate broadened their perspective aimed to measure the educational value of the session.

Furthermore, the debater evaluation included a peer assessment: participants identified the best debater among the group and justified their choice, offered insights into group dynamics and individual performance from multiple perspectives. Overall, the self-evaluation form provided comprehensive feedback on preparation, performance, and educational impact of the debates. This data is valuable for analyzing the results of debates and understanding the factors that contribute to successful debates. Also, it is crucial for participants' development, fostering meta cognition and self-improvement. By analyzing these data, researchers can identify key skills and strategies that distinguish top debaters, offering both theoretical insights and practical guidance for enhancing debate skills in academic and professional contexts.

\subsection{Data Processing and Annotation}

After recording all debate sessions, both automatic and manual post-processing were necessary. This section details the adopted process. The first step was transcribing the recorded audio files, a time-consuming task. To expedite this, we used automatic transcription with free models, testing three machine learning models: wav2vec-large \footnote{\url{https://huggingface.co/facebook/wav2vec2-large}}, whisper-large \footnote{\url{https://huggingface.co/openai/whisper-large-v3}}, and Azure Speech-to-Text \footnote{\url{https://learn.microsoft.com/en-us/azure/ai-services/speech-service/speech-to-text}}. After comparing their performance on the same recordings, we found Azure's model to be the most accurate and suitable for our context. All audio files were then transcribed using Azure, with the outputs exported to CSV files containing the following information:

\begin{itemize}
    \item \textbf{Transcribed text}: The text transcribed from the audio chunk.
    \item \textbf{Start time}: The start time of the transcribed audio chunk.
    \item \textbf{End time}: The end time of the transcribed audio chunk.
\end{itemize}

Errors were manually corrected in the automatic transcriptions. The initial automatic transcription provided a useful starting point, significantly reducing the manual workload. Google Sheets was used as the annotation tool, as it allowed easy editing of text and timestamps. Annotators followed a specific protocol to correct errors in the transcriptions. The primary goal was to ensure accuracy by listening to the audio, comparing it to the automatic transcription, and making necessary corrections. Another task involved handling instances where multiple debaters spoke simultaneously. Although such occurrences were rare due to the debate rules, annotators separated mixed speech into distinct lines, each representing a single speaker.

Also, a final annotation was done to identify the speakers in the audio. A new column was added to the spreadsheet to label each piece of speech with the corresponding debater's identifier. Additionally, if the automatic transcription split speech into smaller fragments due to pauses, annotators combined these fragments into a single line, adjusting the start and end times accordingly. Figure 2 illustrates the entire annotation process for this dataset.

\begin{figure}[ht]
\centering
\includegraphics[width=.8\textwidth]{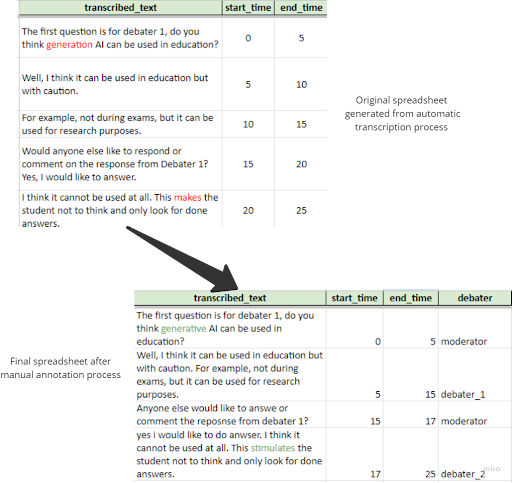}
\caption{Annotation process example}
\label{fig:exampleFig1}
\end{figure}

\subsubsection{DEBISS Corpus Stats}

The proposed corpus was developed to advance the state of the art by introducing a new audio-recorded dataset with the potential to significantly contribute to the scientific study of debates. DEBISS stands out by incorporating distinctive features that differentiate it from existing debate corpora, including semi-structured formats, individual speaker contributions, and spoken discourse. The corpus was designed to fill a gap in the literature regarding non-typical debate structures and aims to support a wide range of NLP applications. Table \ref{tab:debissstats} provides a summary of the corpus stats following the application of the complete methodology.

\begin{table}[h!]
\centering
\caption{DEBISS stats numbers}
\begin{tabular}{@{}ll@{}}
\toprule
\multicolumn{1}{c}{\textbf{Stats}} & \multicolumn{1}{c}{\textbf{Value}} \\ \midrule
Number of debaters & 67 \\
Number of groups & 16 \\
Total audio length of recordings & 9 hours and 37 minutes \\
Number of tokens & 130697 \\
Words lexical diversity\footnotemark & 0.062 \\
\bottomrule
\end{tabular}
\label{tab:debissstats}
\end{table}

 \footnotetext{Measure of vocabulary variation within a text}

\subsubsection{DEBISS Applicability}

As a showcase of the applicability of the DEBISS corpus, we highlight two end-use cases built upon the collected data, named DEBISS-Arg and DEBISS-Eval.

\paragraph{DEBISS-Arg}

Based on the novel DEBISS corpus introduced in this research, a derivative corpus named DEBISS-Arg was developed to advance the state of the art in AM. This fully annotated corpus was created to support the evaluation and of various AM tasks. Specifically, it includes annotations for Argument Discourse Units (ADUs) and non-ADU text segments, as well as for argumentative components such as premises, claims, and evidence. Moreover, the corpus features relation annotations both at the micro level — capturing interactions among components within a single ADU — and at the macro level, which maps relationships across different speech utterances in the debate, thereby capturing argumentive interactions between debaters.

This comprehensive set of labels makes the DEBISS-Arg corpus highly valuable for a wide range of AM tasks. A rigorous methodology for data annotation and processing was followed in the creation of this resource. The full methodological details, including the steps taken to derive DEBISS-Arg from the original DEBISS corpus, are described in \cite{stil} . The Figure\ref{fig:exampleFig1} presents an illustrative example demonstrating how the aforementioned labels are applied within the DEBISS-Arg annotation framework. 

\begin{figure}[ht]
\centering
\includegraphics[width=.9\textwidth]{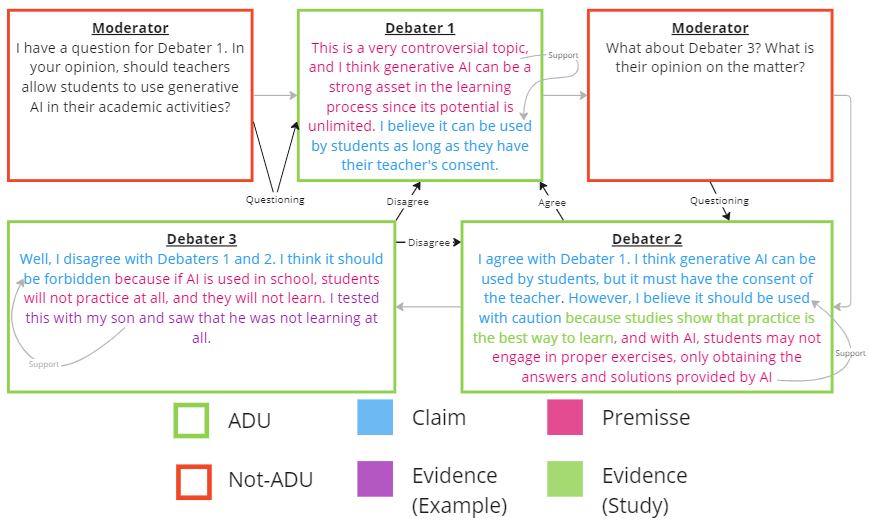}
\caption{Illustrative Example of DEBISS-Arg Data (extracted from [omitted])}
\label{fig:exampleFig1}
\end{figure}

\paragraph{DEBISS-Eval}

DEBISS-Eval\footnote{Avaliable in https://github.com/AINDA-Project-UFCG/DEBISS-Eval}, a specialized subcorpus of the DEBISS academic debate collection, was created to advance research into evaluating debate quality and individual debater performance. A dedicated panel of five expert judges, all with Linguistics degrees and prior debate experience, meticulously evaluated all 16 debates over two weeks. They assessed debaters using a 1-5 Likert scale across established criteria like Organization, Argumentation, Persuasion, and Clarity, yielding rich data including quantitative scores, hard and soft voting results, and extensive qualitative feedback \cite{ericson2011debater}.

\begin{figure}[ht]
\centering
\includegraphics[width=.9\textwidth]{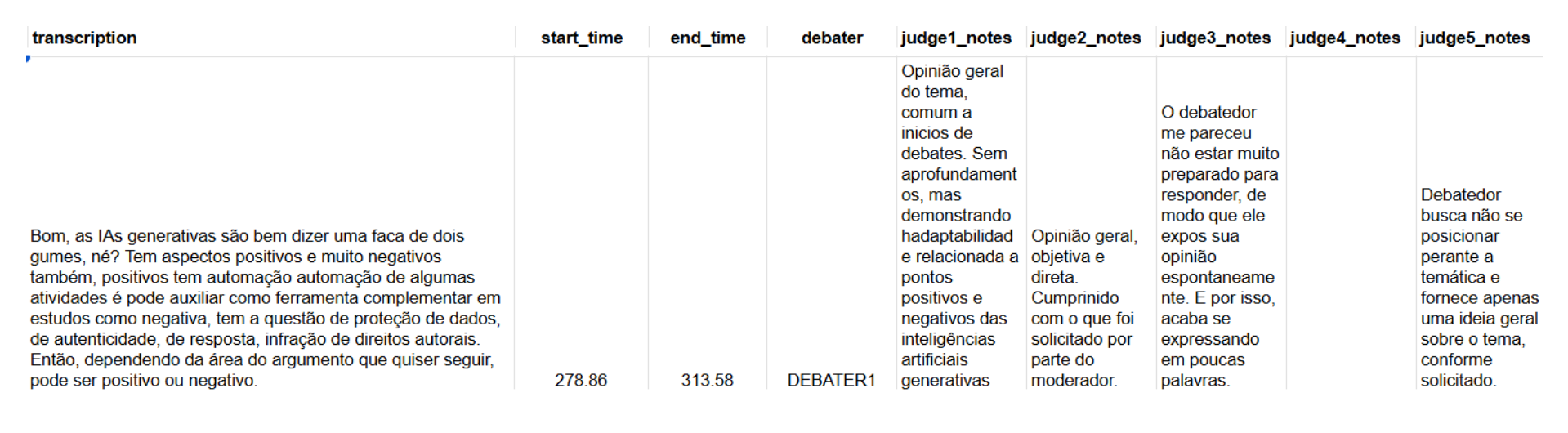}
\caption{Illustrative Example of judges notes on a argument in DEBISS-eval}
\label{fig:judges-notes}
\end{figure}

Moreover, the qualitative data comprises judges' natural language notes on nearly all debater statements and detailed justifications for their overall 'best debater' choices, culminating in 80 detailed evaluation responses. An example of the judges notes can be found at Figure \ref{fig:judges-notes} This wealth of information provides a robust foundation for nuanced analyses of argumentation effectiveness, debate dynamics, and the intricacies of human evaluative judgment in argumentative settings. 

\paragraph{Disfluency Detection}

The DEBISS dataset has also been a focus of studies in text disfluency detection \cite{stil2}. There was an exploration of how Large Language Models (LLMs) perform this task using DEBISS. This research specifically evaluated advanced LLMs, like OpenAI's GPT-4o \cite{openai2024gpt4ocard} to see how well they could identify and remove disfluent elements such as repetitions, hesitations, and corrections from transcribed speech. The study employed various prompt engineering techniques—Zero-Shot, Few-Shot, and Chain-of-Thought prompting—to thoroughly test the LLMs' capabilities with the DEBISS dataset. While some models showed limitations, GPT-4o consistently demonstrated strong performance in both removing disfluencies and maintaining text quality, often outperforming other models. These findings underscore DEBISS's significance as a valuable resource for developing and evaluating cutting-edge solutions in automated disfluency detection.

\section{Conclusion}

The proposed corpus was created to enhance the state of the art by introducing a new audio-recorded corpus with the potential to advance the scientific field of debate analysis. DEBISS goes further by incorporating distinct characteristics that differentiate it from existing debate corpora, such as semi-structured features, individual debates, and spoken content. Also, the contributions goes beyond the data by proposing an methodology for gathering debate in the specifies format

Moreover, this corpus is designed to include rich annotations that support a variety of NLP tasks. These annotations cover speech-to-text transcription, voiceprint identification, speaker diarization, silence detection, disfluency detection, debater quality evaluation (DEBISS-Eval) and argument mining tasks (DEBISS-Arg). As such, the corpus represents a significant advancement in the availability of open data for NLP research in Brazilian Portuguese.

Although the dataset presents a well-defined methodology and comprehensive stats numbers, it is important to highlight some key limitations. One such limitation is the narrow thematic scope of the debates, which may affect the generalizability of the dataset. Expanding the range of debate topics and incorporating a broader thematic diversity would enhance its applicability. Additionally, collecting data from more heterogeneous participant groups — such as students from varied educational backgrounds and levels — would further strengthen the dataset’s representativeness. These limitations can be addressed in future work by applying the proposed methodology to a wider array of themes and more diverse debater profiles.

\bibliographystyle{sbc}
\bibliography{sbc-template}

@article{vrana2017saying,
  title={Saying whatever it takes: Creating and analyzing corpora from US presidential debate transcripts},
  author={Vrana, Leo and Schneider, Gerold},
  year={2017},
  publisher={University of Zurich}
}

@inproceedings{de2018polly,
  title={The Polly corpus: Online political debate in Germany},
  author={De Smedt, Tom and Jaki, Sylvia},
  booktitle={of the 6th Conference on Computer-Mediated Communication (CMC) and Social Media Corpora (CMC-corpora 2018)},
  pages={33},
  year={2018}
}

@inproceedings{carvalho2011liars,
  title={Liars and saviors in a sentiment annotated corpus of comments to political debates},
  author={Carvalho, Paula and Sarmento, Lu{\'\i}s and Teixeira, Jorge and Silva, M{\'a}rio J},
  booktitle={Proceedings of the 49th Annual Meeting of the Association for Computational Linguistics: Human Language Technologies},
  pages={564--568},
  year={2011}
}

@inproceedings{hautli2022qt30,
  title={Qt30: A corpus of argument and conflict in broadcast debate},
  author={Hautli-Janisz, Annette and Kikteva, Zlata and Siskou, Wassiliki and Gorska, Kamila and Becker, Ray and Reed, Chris},
  booktitle={Proceedings of the 13th Language Resources and Evaluation Conference},
  pages={3291--3300},
  year={2022},
  organization={European Language Resources Association (ELRA)}
}

@inproceedings{abbott2016internet,
  title={Internet argument corpus 2.0: An sql schema for dialogic social media and the corpora to go with it},
  author={Abbott, Rob and Ecker, Brian and Anand, Pranav and Walker, Marilyn},
  booktitle={Proceedings of the Tenth International Conference on Language Resources and Evaluation (LREC'16)},
  pages={4445--4452},
  year={2016}
}

@article{durmus2019corpus,
  title={A corpus for modeling user and language effects in argumentation on online debating},
  author={Durmus, Esin and Cardie, Claire},
  journal={arXiv preprint arXiv:1906.11310},
  year={2019}
}

@article{khodak2017large,
  title={A large self-annotated corpus for sarcasm},
  author={Khodak, Mikhail and Saunshi, Nikunj and Vodrahalli, Kiran},
  journal={arXiv preprint arXiv:1704.05579},
  year={2017}
}

@article{stranisci2021expression,
  title={The expression of moral values in the twitter debate: a corpus of conversations},
  author={Stranisci, Marco and De Leonardis, Michele and Bosco, Cristina and Patti, Viviana},
  journal={IJCoL. Italian Journal of Computational Linguistics},
  volume={7},
  number={7-1, 2},
  pages={113--132},
  year={2021},
  publisher={Accademia University Press}
}

@inproceedings{lai2018stance,
  title={Stance evolution and twitter interactions in an italian political debate},
  author={Lai, Mirko and Patti, Viviana and Ruffo, Giancarlo and Rosso, Paolo},
  booktitle={Natural Language Processing and Information Systems: 23rd International Conference on Applications of Natural Language to Information Systems, NLDB 2018, Paris, France, June 13-15, 2018, Proceedings 23},
  pages={15--27},
  year={2018},
  organization={Springer}
}

@article{ruiz2021vivesdebate,
  title={Vivesdebate: A new annotated multilingual corpus of argumentation in a debate tournament},
  author={Ruiz-Dolz, Ramon and Nofre, Montserrat and Taul{\'e}, Mariona and Heras, Stella and Garc{\'\i}a-Fornes, Ana},
  journal={Applied Sciences},
  volume={11},
  number={15},
  pages={7160},
  year={2021},
  publisher={MDPI}
}

@inbook{0eac738616614094950bb74635ce3d49,
title = "Mining Ethos in Political Debate",
author = "Rory Duthie and Katarzyna Budzynska and Chris Reed",
note = "This research was supported in part by EPSRC in the UK under grant EP/M506497/1 and in part by the Polish National Science Centre under grant 2015/18/M/HS1/00620.",
year = "2016",
doi = "10.3233/978-1-61499-686-6-299",
language = "English",
isbn = "9781614996859",
volume = "287",
series = "Frontiers in Artificial Intelligence and Applications",
publisher = "IOS Press",
pages = "299--310",
editor = "Pietro Baroni and Gordon, {Thomas F.} and Tatjana Scheffler and Manfred Stede",
booktitle = "Computational Models of Argument",
address = "Netherlands",
}

@inproceedings{mestre-etal-2021-arg,
    title = "{M}-Arg: Multimodal Argument Mining Dataset for Political Debates with Audio and Transcripts",
    author = "Mestre, Rafael  and
      Milicin, Razvan  and
      Middleton, Stuart E.  and
      Ryan, Matt  and
      Zhu, Jiatong  and
      Norman, Timothy J.",
    editor = "Al-Khatib, Khalid  and
      Hou, Yufang  and
      Stede, Manfred",
    booktitle = "Proceedings of the 8th Workshop on Argument Mining",
    month = nov,
    year = "2021",
    address = "Punta Cana, Dominican Republic",
    publisher = "Association for Computational Linguistics",
    url = "https://aclanthology.org/2021.argmining-1.8",
    doi = "10.18653/v1/2021.argmining-1.8",
    pages = "78--88",
}

@inproceedings{mancini-etal-2022-multimodal,
    title = "Multimodal Argument Mining: A Case Study in Political Debates",
    author = "Mancini, Eleonora  and
      Ruggeri, Federico  and
      Galassi, Andrea  and
      Torroni, Paolo",
    editor = "Lapesa, Gabriella  and
      Schneider, Jodi  and
      Jo, Yohan  and
      Saha, Sougata",
    booktitle = "Proceedings of the 9th Workshop on Argument Mining",
    month = oct,
    year = "2022",
    address = "Online and in Gyeongju, Republic of Korea",
    publisher = "International Conference on Computational Linguistics",
    url = "https://aclanthology.org/2022.argmining-1.15",
    pages = "158--170",
}

@inproceedings{brasnam,
 author = {João Pedro Sousa and Rodrigo Nascimento and Renata Araujo and Orlando Coelho},
 title = { Não se perca no debate! Mineração de Argumentação em Redes Sociais},
 booktitle = {Anais do X Brazilian Workshop on Social Network Analysis and Mining},
 location = {Evento Online},
 year = {2021},
 issn = {2595-6094},
 pages = {139--150},
 publisher = {SBC},
 address = {Porto Alegre, RS, Brasil},
 doi = {10.5753/brasnam.2021.16132},
 url = {https://sol.sbc.org.br/index.php/brasnam/article/view/16132}
}

@inproceedings{habernal-gurevych-2016-argument,
    title = "Which argument is more convincing? Analyzing and predicting convincingness of Web arguments using bidirectional {LSTM}",
    author = "Habernal, Ivan  and
      Gurevych, Iryna",
    editor = "Erk, Katrin  and
      Smith, Noah A.",
    booktitle = "Proceedings of the 54th Annual Meeting of the Association for Computational Linguistics (Volume 1: Long Papers)",
    month = aug,
    year = "2016",
    address = "Berlin, Germany",
    publisher = "Association for Computational Linguistics",
    url = "https://aclanthology.org/P16-1150",
    doi = "10.18653/v1/P16-1150",
    pages = "1589--1599",
}

@inproceedings{boltuzic-snajder-2016-fill,
    title = "Fill the Gap! Analyzing Implicit Premises between Claims from Online Debates",
    author = "Boltu{\v{z}}i{\'c}, Filip  and
      {\v{S}}najder, Jan",
    editor = "Reed, Chris",
    booktitle = "Proceedings of the Third Workshop on Argument Mining ({A}rg{M}ining2016)",
    month = aug,
    year = "2016",
    address = "Berlin, Germany",
    publisher = "Association for Computational Linguistics",
    url = "https://aclanthology.org/W16-2815",
    doi = "10.18653/v1/W16-2815",
    pages = "124--133",
}

@inproceedings{chakrabarty-etal-2019-ampersand,
    title = "{AMPERSAND}: Argument Mining for {PERS}u{A}sive o{N}line Discussions",
    author = "Chakrabarty, Tuhin  and
      Hidey, Christopher  and
      Muresan, Smaranda  and
      McKeown, Kathy  and
      Hwang, Alyssa",
    editor = "Inui, Kentaro  and
      Jiang, Jing  and
      Ng, Vincent  and
      Wan, Xiaojun",
    booktitle = "Proceedings of the 2019 Conference on Empirical Methods in Natural Language Processing and the 9th International Joint Conference on Natural Language Processing (EMNLP-IJCNLP)",
    month = nov,
    year = "2019",
    address = "Hong Kong, China",
    publisher = "Association for Computational Linguistics",
    url = "https://aclanthology.org/D19-1291",
    doi = "10.18653/v1/D19-1291",
    pages = "2933--2943",
}

@misc{ericson2011debater,
  title={The debater’s guide},
  author={Ericson, JM},
  year={2011},
  publisher={Southern Illinois University Press}
}

@misc{openai2024gpt4ocard,
      title={GPT-4o System Card}, 
      author={OpenAI and : and Aaron Hurst and Adam Lerer and Adam P. Goucher and Adam Perelman and Aditya Ramesh et. al},
      year={2024},
      eprint={2410.21276},
      archivePrefix={arXiv},
      primaryClass={cs.CL},
      url={https://arxiv.org/abs/2410.21276}, 
}

@inproceedings{stil,
 author = {David Pereira and Daniela Simão and Claudio Campelo},
 title = { DEBISS-Arg: An In Depth Data Annotation Protocol and Corpus for Argument Mining in Semi Structured Debates},
 booktitle = {Anais do XVI Simpósio Brasileiro de Tecnologia da Informação e da Linguagem Humana},
 location = {Fortaleza/CE},
 year = {2025},
 issn = {0000-0000},
 pages = {334--348},
 publisher = {SBC},
 address = {Porto Alegre, RS, Brasil},
 doi = {10.5753/stil.2025.37836},
 url = {https://sol.sbc.org.br/index.php/stil/article/view/37836}
}

@inproceedings{stil2,
 author = {Pedro L. de Lima and Cláudio E. Campelo},
 title = { Disfluency Detection and Removal in Speech Transcriptions via Large Language Models},
 booktitle = {Anais do XV Simpósio Brasileiro de Tecnologia da Informação e da Linguagem Humana},
 location = {Belém/PA},
 year = {2024},
 issn = {0000-0000},
 pages = {227--235},
 publisher = {SBC},
 address = {Porto Alegre, RS, Brasil},
 doi = {10.5753/stil.2024.245417},
 url = {https://sol.sbc.org.br/index.php/stil/article/view/31135}
}

\end{document}